\documentclass[10pt,twocolumn,letterpaper]{article}

\usepackage[pagenumbers]{cvpr}

\usepackage{fvextra}
\usepackage[most]{tcolorbox}
\usepackage{booktabs}
\usepackage{multirow}
\usepackage{cuted}

\def \customparskip {.3em}
\renewcommand{\paragraph}[1]{\vspace{\customparskip}\noindent\textbf{#1}}

\newcommand{\supp}[0]{\texttt{Supplementary Material}\xspace}
\newcommand{\visualsync}[0]{VisualSync~\cite{liu2025visualsync}\xspace}
\newcommand{\syncnerf}[0]{Sync-NeRF~\cite{kim2024syncnerf}\xspace}
\newcommand{\mastr}[0]{MASt3R~\cite{leroy2024mast3r}\xspace}

\newcommand{\vggt}[0]{VGGT~\cite{wang2025vggt}\xspace}
\newcommand{\cotracker}[0]{CoTracker3~\cite{karaev2025cotracker3}\xspace}
\newcommand{\vda}[0]{Video Depth Anything~\cite{chen2025videodepthanything}\xspace}
\newcommand{\velodepth}[0]{VeloDepth~\cite{piccinelli2026velodepth}\xspace}
\newcommand{\samthree}[0]{SAM~3~\cite{carion2025sam3}\xspace}
\newcommand{\gpt}[0]{GPT-5.4~\cite{openai2026gpt54}\xspace}
\newcommand{\unifourd}[0]{Uni4D~\cite{yao2025uni4d}\xspace}
\newcommand{\sesyn}[0]{SeSyn-Net~\cite{yin2022sesyn}\xspace}
\newcommand{\poseagg}[0]{Human Pose Aggregation~\cite{delattre2026poseaggregation}\xspace}
\newcommand{\hrnet}[0]{HRNet~\cite{sun2019hrnet}\xspace}

\newcommand{\egohumans}[0]{EgoHumans~\cite{khirodkar2023egohumans}\xspace}
\newcommand{\panoptic}[0]{CMU Panoptic~\cite{joo2015panoptic}\xspace}
\newcommand{\threedpop}[0]{3D-POP~\cite{naik2023threedpop}\xspace}
\newcommand{\udbd}[0]{UDBD~\cite{kim2024syncnerf}\xspace}

\definecolor{cvprblue}{rgb}{0.21,0.49,0.74}
\usepackage[pagebackref,breaklinks,colorlinks,allcolors=cvprblue]{hyperref}
\usepackage{amsmath}
\usepackage{amssymb}
\usepackage{booktabs}
\usepackage{multirow}

\title{Synchronizing Cameras via 3D Object Relationships}

\author{
Sujong Shin$^{1}$ \quad
Sang-Il Choi$^{1}$ \quad
Kwang Moo Yi$^{2}$\\[0.25cm]
{\normalfont\normalsize
$^{1}$Dankook University \quad
$^{2}$University of British Columbia
}\\[0.05cm]
{\tt\small shinsjn@dankook.ac.kr \quad choisi@dankook.ac.kr \quad kmyi@cs.ubc.ca}
}

\begin{document}
\maketitle
\begin{strip}
    \centering
    \begin{minipage}[c]{0.66\textwidth}
        \includegraphics[width=\linewidth]{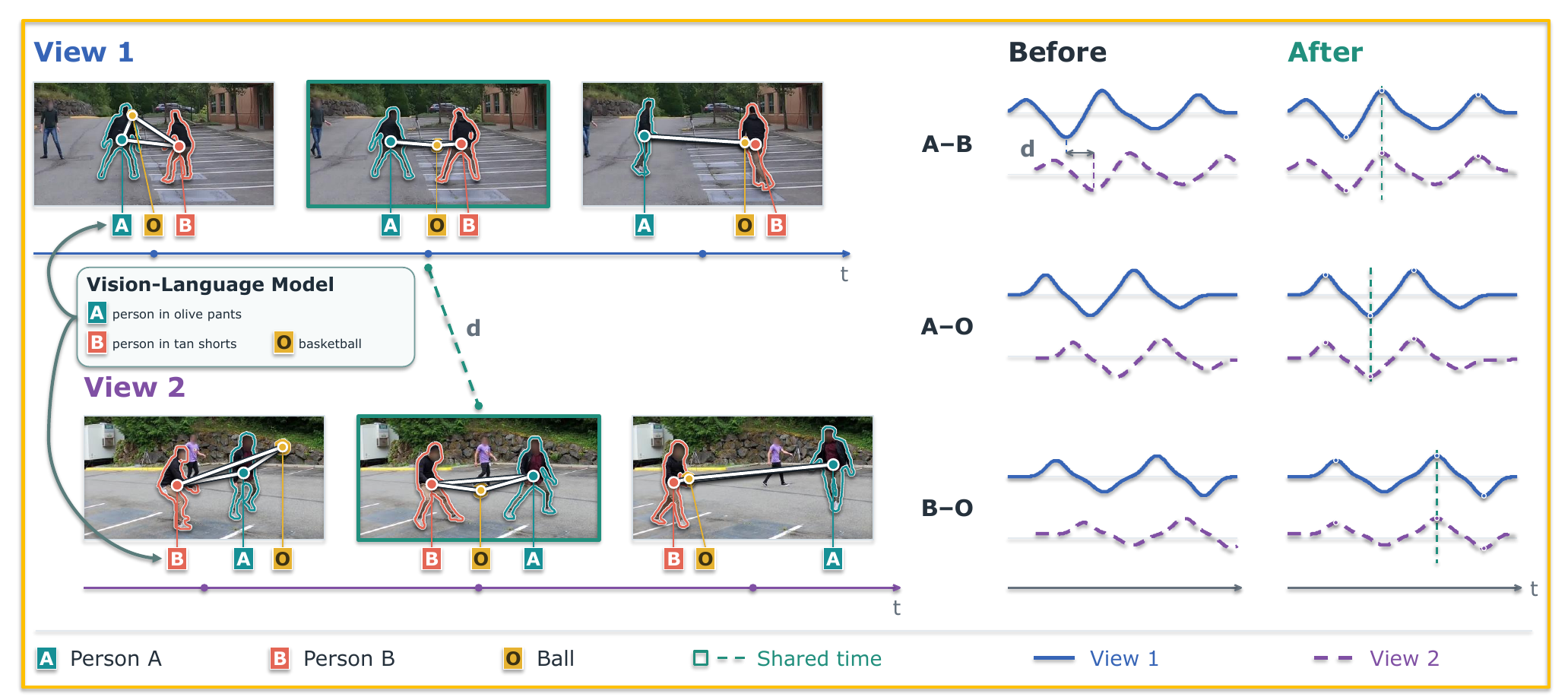}
    \end{minipage}\hspace{0.02\textwidth}%
    \begin{minipage}[c]{0.30\textwidth}
        \includegraphics[width=\linewidth]{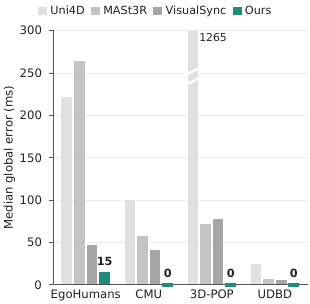}
    \end{minipage}
    \captionsetup{hypcap=false}
    \captionof{figure}{
        \textbf{Synchronization through object relationships.}
        (Left) With a vision-language model, we identify the same objects in every view, here persons A and B and the ball O.
        We then measure the distances between them in 3D within each view.
        These distances agree across views up to scale once we shift the videos by the correct offset~$d$, without matching points across views.
        (Right) We plot the synchronization error in terms of median global error of our method and other methods on four standard datasets.
        Our method provides significantly better synchronization.
    }
    \captionsetup{hypcap=true}
    \label{fig:teaser}
\end{strip}

\begin{abstract}
In this paper, we synchronize multiple videos of the same dynamic event through object relationships.
Driven by dynamic reconstruction and novel view synthesis, many recent synchronization methods rely on pixel-level evidence, such as point correspondences across views or the photometric losses of a reconstructed scene.
In dynamic scenes, however, fast motion and occlusions make the moving regions that carry much of the timing information hard to match across views.
We thus propose to bring back the traditional wisdom that objects are reliable cues for global reasoning about cameras.
Measured in 3D, the relationships between objects, such as the distance between two people, are stable across all views up to scale---they can serve as cues for alignment.
To compare these relationships across cameras, we only need the identity of each object in every view, as the cues are relational.
We obtain these identities by giving a vision-language model a few frames of every camera and prompting it to identify the objects in the scene.
We then segment and track each identified object through every video with an open-vocabulary segmentor.
To obtain the 3D locations of objects, we use a pretrained depth estimator.
Where depth or segmentation fails, we complement these 3D cues with 2D motion from point tracks.
Finally, we synchronize the cameras by finding the temporal offsets at which the cues of different cameras agree best, first for each camera pair and then for all cameras jointly with robust optimization.
We evaluate our method on four standard benchmarks, outperforming the state of the art.
\end{abstract}

\section{Introduction}
\label{sec:intro}

Many computer vision tasks, such as dynamic 3D reconstruction~\cite{luiten2024dynamic3dgaussians}, free-viewpoint rendering~\cite{li2022neural3dvideo}, and multi-view motion analysis~\cite{iskakov2019learnable}, rely on temporally aligned observations.
However, independently recorded videos may start at different times and lack shared hardware triggers, timecodes, or audio cues~\cite{shrestha2007audio,liu2025visualsync}.
Misaligned videos then provide inconsistent observations of the same moment, which harms dynamic reconstruction~\cite{vo2016spatiotemporal,kim2024syncnerf,choi2025humans}.
It is thus important to synchronize such RGB videos after capture.

Various synchronization methods~\cite{caspi2002spatiotemporal,wolf2006widebaseline,pundik2010epipolar,albl2017twoview,yin2022sesyn,naaman2025multiple,kim2024syncnerf,xu2026dynamicgaussian,liu2025visualsync} have been proposed.
Driven by dynamic reconstruction and novel view synthesis, many recent methods rely on pixel-level evidence.
Some estimate the temporal offsets of the videos jointly with a radiance field or a set of 3D Gaussians~\cite{kim2024syncnerf,xu2026dynamicgaussian,lee2026synctrack4d}, and others match point trajectories across views~\cite{liu2025visualsync}.
But properly estimating the motion of each pixel in itself is a difficult problem.
In dynamic scenes, the moving regions carry much of the timing information, yet their fast motion and frequent occlusions~\cite{butler2012sintel} make them hard to match across views~\cite{zhang2025monst3r}.
D$^2$USt3R~\cite[Fig.~2]{han2025d2ust3r} and VisualSync~\cite[Fig.~12]{liu2025visualsync} show such matching failures on moving objects.

We instead return to the traditional wisdom of using objects as reliable cues for global reasoning about cameras.
Early synchronization methods aligned cameras through tracked objects, their trajectories, and their silhouettes~\cite{stein1999tracking,lee2000monitoring,caspi2006featurebased,sinha2010silhouettes}.
People and objects have also long served to calibrate and localize cameras~\cite{lv2006walking,hoiem2006perspective,salasmoreno2013slampp,yang2019cubeslam}.
In each camera, an object can be found and measured more reliably than a pixel, as the object pools the evidence of many pixels.
What these traditional ideas lack is a proper modernization.
Modern foundation models provide it, as they can identify~\cite{openai2026gpt54}, segment~\cite{carion2025sam3}, and locate in 3D~\cite{chen2025videodepthanything} the objects of a scene.

We thus propose to synchronize cameras by comparing how objects move and relate to each other in 3D, which we can measure in each camera independently.
Up to scale and perception errors, the distance between two people or the speed of a ball in 3D does not depend on where a static camera stands.
These cues thus agree across cameras, once we align the cameras in time as shown in \cref{fig:teaser}.
In other words, we only need the identities of the objects across views, without needing to match points.

We obtain the identities of the objects across views by giving a vision-language model (VLM)~\cite{openai2026gpt54} a few frames of every camera and prompting it for a short description of each object.
We then segment and track each identified object through every video with an open-vocabulary segmentor~\cite{carion2025sam3}.
When the VLM fails to distinguish instances, such as by assigning a single identity to multiple pigeons of a flock, we match them across views by how their 3D distances to the other objects change over time.
As the descriptions are at the object level, the identities can be of any category, and our cues are not limited to people as in other existing work~\cite{yin2022sesyn,delattre2026poseaggregation}.

To obtain the 3D locations of the objects, we use the pretrained metric \vda and the ground-truth intrinsics of each camera.
These 3D cues, however, can become unreliable when depth or segmentation fails, as for small, textureless, or occluded objects.
We thus complement them with 2D motion from point tracks within each view.
With these cues, we optimize for the best match across cues.
As our ablation in \cref{sec:cue-ablation} shows, each cue alone fails on at least one dataset, while all cues together give the state-of-the-art results overall.

To synchronize the cameras with these cues, we follow the two steps of \visualsync, but tailor both to object cues.
We first estimate the offset of each camera pair as the one at which their cues agree best.
To compare the cues of two cameras regardless of the scale of their depth estimates, we correlate them.
We then combine the pairwise offsets into one consistent offset per camera with a robust optimization, in which we weight each pair by its confidence.

To summarize, our contributions are:
\begin{itemize}
    \item we bring object-level cues back to camera synchronization, by comparing the 3D relationships between objects across cameras;

    \item we obtain the identities of objects across views with foundation models, instead of matching points;

    \item we use the 3D distances between objects as cues, and compare them across cameras by correlation, regardless of the scale of each camera's depth; and

    \item we outperform the state of the art in camera synchronization on four standard benchmarks.
\end{itemize}

\section{Related work}
\label{sec:related}

We review visual methods for synchronization, and organize them by the level at which they compare the videos, from pixels and points to objects.
We then review the use of objects in camera calibration and localization, and finally the foundation models that we use to measure the objects in each video.

\paragraph{Pixel- and point-level synchronization.}
Many synchronization methods compare the videos at the level of pixels or points.
Some compare features of whole frames, such as signatures of image change over time~\cite{yan2004videosynchronization,ushizaki2006appearance,alnuaimi2012concor,pereira2017videosynchronization}, self-similarity matrices~\cite{dexter2009multiview}, or learned frame features~\cite{naaman2025multiple}.
Features of whole frames~\cite{yan2004videosynchronization,dexter2009multiview,naaman2025multiple}, however, differ between cameras when an object is visible in only some of them~\cite{ushizaki2006appearance}.
We use one such feature, the 2D motion of tracked points, only to complement our object cues.

Other methods relate pixels across views through homographies~\cite{caspi2002spatiotemporal} or along epipolar lines~\cite{pundik2010epipolar}, points through epipolar geometry~\cite{albl2017twoview} or bundle adjustment~\cite{vo2016spatiotemporal}, and point trajectories through rank constraints~\cite{rao2003viewinvariant,wolf2006widebaseline,tresadern2009human} or their 3D reconstruction~\cite{wang2017trajectory}.
Recent methods relate them through a shared reconstruction, by estimating the offsets jointly with a dynamic radiance field~\cite{kim2024syncnerf} or dynamic 3D Gaussians~\cite{xu2026dynamicgaussian,lee2026synctrack4d}.
\visualsync matches point trajectories of the moving objects across views, and finds the offset of each camera pair that minimizes the epipolar errors of these matches.
Such relations between pixels or points, however, can fail on moving objects~\cite[Fig.~2]{han2025d2ust3r},~\cite[Fig.~12]{liu2025visualsync}.
We instead compare the objects through cues that we compute in each camera independently.

\paragraph{Object-level synchronization.}
At the level of objects, early methods relate the trajectories of tracked objects~\cite{stein1999tracking,lee2000monitoring,caspi2006featurebased,wang2016trajectorypulse} or their silhouettes~\cite{sinha2010silhouettes} across views.
Other methods compare human poses~\cite{takahashi2018humanpose,wu2019pose,sun2020motionrhythms,yin2022sesyn,liu2024fractional,delattre2026poseaggregation}, or initialize the offsets of a dynamic reconstruction from the estimated shape and pose of people~\cite{choi2025humans}.
Some early methods, however, must estimate the geometry between the views together with the offsets, as a homography~\cite{stein1999tracking,lee2000monitoring,caspi2006featurebased} or epipolar geometry~\cite{caspi2006featurebased,sinha2010silhouettes}.
Their offsets are thus only as reliable as this geometry, which they estimate from 2D tracks or silhouettes alone.
The pose-based methods, in turn, apply only to people.
We follow the tradition of object-level synchronization, but compare \emph{3D cues of each camera} in its own coordinate system and align them.
Our objects can also be of any category that a short description can name.

Among the pose-based methods, those that compare \emph{view-invariant} pose features~\cite{wu2019pose,yin2022sesyn,delattre2026poseaggregation} are the closest in spirit to ours.
\poseagg handles several people without identifying them across views, by pooling their poses into one descriptor per frame.
We instead identify the objects across views with a VLM, and match the instances that it cannot tell apart through their 3D relationships to the other objects.

\paragraph{Objects in camera calibration and localization.}
Besides synchronization, objects have long been a cue for other global problems of camera geometry.
Walking people have been used to calibrate cameras~\cite{lv2006walking,krahnstoever2005bayesian}, detected objects to estimate the camera viewpoint~\cite{hoiem2006perspective}, and objects as landmarks to localize cameras in simultaneous localization and mapping~\cite{salasmoreno2013slampp,nicholson2019quadricslam,yang2019cubeslam}.
Objects have also been used to relocalize cameras from very different viewpoints~\cite{zins2022oaslam}, and semantics to localize them under changes of viewpoint and appearance~\cite{schonberger2018semantic}.
While these methods place cameras in space, we use objects to align cameras in time.

\paragraph{Foundation models for objects.}
To measure the objects in each video, we build on foundation models that need no training on the scene.
VLMs can describe the objects in an image in text~\cite{openai2026gpt54}.
\samthree segments and tracks through a video every instance of a concept that a noun phrase describes.
Depth estimators predict metric depth from a single image~\cite{bochkovskii2025depthpro} and consistent depth over a video~\cite{chen2025videodepthanything,piccinelli2026velodepth}.
\visualsync also uses a VLM and open-vocabulary segmentation to find the moving objects across views, but only to restrict its point matches to these objects.
We instead locate the objects in 3D, so that we can measure the relationships between them in each camera independently.

\section{Method}
\label{sec:method}

\begin{figure*}[t]
\centering
\includegraphics[width=0.95\linewidth,keepaspectratio]{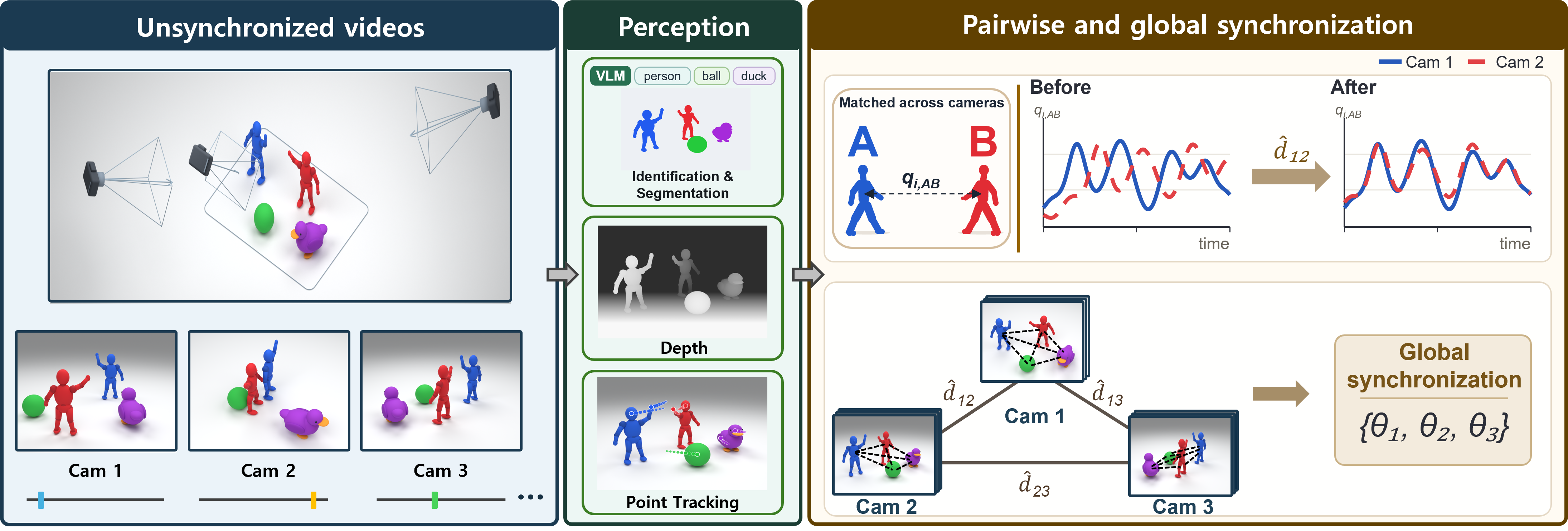}
\caption{\textbf{Overview of our method.}
To identify the objects consistently across views, we give a vision-language model (VLM) frames of all unsynchronized videos at once, and get a short description of each object.
To measure how the objects move and how far apart they are in 3D, we segment them in every video and combine the masks with monocular depth estimates.
When the VLM cannot tell instances apart, such as the two persons A and B, we match them across views by their 3D distances to the other objects.
As a fallback for when depth or segmentation fails, we also track points to measure the 2D motion in each video.
With these cues, we align each pair of videos, and then all videos together.
We thus synchronize the cameras through the 3D relationships between objects, without matching points across views.
}
\label{fig:method-overview}
\end{figure*}

Our core idea is that, up to scale, objects relate to each other in 3D in the same way in every view.
We thus synchronize the cameras by finding the offsets at which these relationships, measured within each view, agree.
\Cref{fig:method-overview} shows our pipeline.
In \cref{sec:problem-formulation}, we first write this agreement as an objective, assuming that the objects are identified across views and located in 3D.
In \cref{sec:object-identities}, we then obtain these identities and 3D locations, and in \cref{sec:pairwise-alignment}, we finally solve for the offsets.

\subsection{Problem formulation}
\label{sec:problem-formulation}

Let us first formalize the problem.
Our input is $N$ RGB videos of the same dynamic event, with known camera intrinsics $\{K_i\}_{i=1}^{N}$.
We further assume that all cameras share one frame rate~\cite{albl2017twoview}.
We then estimate one temporal offset $\theta_i$ per camera, such that frame $t$ of camera $i$ shows the event at global time $t+\theta_i$.
Frame $t$ of camera $i$ and frame $t+d$ of camera $j$ thus show the same moment when $\theta_i-\theta_j=d$.
Without loss of generality, we set camera 1 as the reference camera, that is, we fix $\theta_1=0$.
We find these offsets through the objects that the cameras observe.
For the sake of easy explanation, let us for now assume that the objects are identified across each pair of views and located in 3D (\cref{sec:object-identities}).

\paragraph{Synchronizing by agreement.}
To find the offsets, we compare the cues that we compute about the objects in each camera independently, such as the speed of an object.
As the objects relate to each other in 3D in the same way in every view, these cues agree across cameras up to scale once the cameras are aligned in time.
We thus seek the offsets at which these cues agree best across cameras.
We score the agreement of cue $c$ between cameras $i$ and $j$ at offset $d$ by $r_{ij,c}(d)$, and maximize it over all camera pairs and cues,
\begin{equation}
\operatorname*{arg\,max}_{\boldsymbol{\theta},\,\theta_1=0}\ \sum_{i<j}\ \sum_{c\in\mathcal{C}_{ij}}\ r_{ij,c}(\theta_i-\theta_j),
\label{eq:objective}
\end{equation}
where $\boldsymbol{\theta}=(\theta_1,\ldots,\theta_N)$, and $\mathcal{C}_{ij}$ is the set of cues available in both cameras $i$ and $j$.
We next define the cues and this score.

\paragraph{Motion and relationship cues.}
\label{sec:per-view-cues}
Given the 3D point $\mathbf{p}_{i,o}(t)$ of each object $o$ in camera $i$ at frame $t$, we compute the displacement and the speed of each object, and the distance between each pair of objects $a$ and $b$,
\begin{equation}
\begin{aligned}
\Delta\mathbf{p}_{i,o}(t) &= \mathbf{p}_{i,o}(t)-\mathbf{p}_{i,o}(t-1),\\
\qquad v_{i,o}(t) &= \lVert \Delta\mathbf{p}_{i,o}(t) \rVert_2,\\
q_{i,ab}(t) &= \lVert \mathbf{p}_{i,a}(t)-\mathbf{p}_{i,b}(t) \rVert_2.
\end{aligned}
\label{eq:object-cues}
\end{equation}
These cues are thus in the coordinates of each camera, with its own depth scale.

These 3D cues, however, rely on depth and segmentation.
To keep a cue where either fails, we also track points within each video with \cotracker, and take the mean magnitude of their 2D displacements at frame $t$ as the 2D motion $v^{\mathrm{2D}}_i(t)$ (see \supp).

\paragraph{Scale-invariant alignment.}
If the depth of each camera $i$ is correct up to a scale $\alpha_i$ that is constant over time (see \supp), the speeds in two static cameras agree up to scale once the cameras are aligned in time, without the need for camera alignment,
\begin{equation}
v_{i,o}(t) \approx \frac{\alpha_i}{\alpha_j}\, v_{j,o}(t+\theta_i-\theta_j),
\label{eq:invariance}
\end{equation}
and likewise for the distance $q_{i,ab}$.
On a moving camera, such as an egocentric one, the camera motion adds to the speed, but the distance still stays invariant.
The displacement $\Delta\mathbf{p}_{i,o}$ and the 2D motion $v^{\mathrm{2D}}_i$, in contrast, depend on the viewpoint even on static cameras.
They still change at the same moments in every view, however, for instance when an object starts or stops moving.

To compare how the cues rise and fall over time regardless of the scale $\alpha_i/\alpha_j$ in \cref{eq:invariance}, we follow correlation-based synchronization~\cite{schweiger2011concor,alnuaimi2012concor} and correlate them.
With the cues of \cref{eq:object-cues} and the 2D motion, we write the values $y_{i,c}(t)$ of every cue $c$ in camera $i$ at frame $t$ as the set
\begin{equation}
\begin{aligned}
\{y_{i,c}(t)\}_{c} = {}& \{v_{i,o}(t)\}_{o} \cup \{q_{i,ab}(t)\}_{a<b} \cup \{v^{\mathrm{2D}}_i(t)\}\\
&\cup \{\Delta p^{x}_{i,o}(t),\, \Delta p^{y}_{i,o}(t),\, \Delta p^{z}_{i,o}(t)\}_{o},
\end{aligned}
\label{eq:cue-set}
\end{equation}
where $\Delta p^{x}_{i,o}$, $\Delta p^{y}_{i,o}$, and $\Delta p^{z}_{i,o}$ are the coordinates of $\Delta\mathbf{p}_{i,o}$.
We then write the score as
\begin{equation}
r_{ij,c}(d) = \operatorname{corr}_t\bigl(y_{i,c}(t),\, y_{j,c}(t+d)\bigr),
\label{eq:correlation}
\end{equation}
where $\operatorname{corr}_t$ is the Pearson correlation over the frames $t$ at which the cue is available in both cameras.
For the speed and the distance, this correlation thus ideally peaks at $d=\theta_i-\theta_j$ (\cref{eq:invariance}).
Our ablation in \cref{sec:cue-ablation} shows that adding the displacement and the 2D motion to the speed and the distance still lowers our error on every dataset.
In \cref{sec:pairwise-alignment}, we solve \cref{eq:objective} approximately, in two steps, after we obtain the identities and 3D locations of the objects.

\subsection{Object identities and 3D locations}
\label{sec:object-identities}

To obtain the identities and 3D locations of the objects, we identify, segment, and locate the objects with off-the-shelf methods, and then match across views the instances that remain ambiguous.

\paragraph{Identifying the objects with a VLM.}
To identify the objects, we prompt the VLM \gpt once, with the first, middle, and last frames of every camera, so that it sees which objects move.
In this single prompt, we instruct it to find the objects in each camera, merge the same object across cameras, and select the objects whose motion is informative for synchronization (see \supp).
It outputs the selected objects as a JSON list, with a \emph{description} of each object, a short noun phrase such as ``man in a pink checkered shirt''.
As this one list covers all cameras, each object has the same description in every camera.
A description, however, cannot tell apart instances that look alike, such as the pigeons of a flock.
We thus match such instances across views later, by their 3D distances to the other objects.

\paragraph{Locating the objects in 3D.}
To locate each identified object in 3D, we first segment and track through each video every instance that fits each description, by giving the description to \samthree as a text prompt.
We then obtain metric depth from the large metric model of \vda, and back-project~\cite{hartley2004multipleview} the centroid of each mask with $K_i$ and the median depth of the mask, which gives $\mathbf{p}_{i,o}(t)$.
When a description fits only one instance, such as a single ball, the description already identifies this instance in every camera, so we can compare its cues across cameras right away.
When several instances share a description, however, we do not know which instance in one camera is which in another.

\paragraph{Matching instances that share a description.}
To compare across cameras the cues of the instances that share a description, we first match these instances across views, one to one for each camera pair.
When the two cameras see different numbers of them, we match only as many as the smaller number, keeping the pairs with the highest total score in \cref{eq:instance-search}.
We represent each instance $a$ of camera $i$ by its distances to the other objects, which we sort in ascending order so that the feature is invariant to the order of the objects,
\begin{equation}
h_{i,a}(t) = \operatorname*{sort}_{o\neq a}\, \lVert \mathbf{p}_{i,a}(t)-\mathbf{p}_{i,o}(t) \rVert_2.
\label{eq:instance-descriptor-main}
\end{equation}
If instance $a$ of camera $i$ and instance $b$ of camera $j$ are the same, their distances to the other objects rise and fall together once the cameras are aligned in time.
We thus score how well $a$ and $b$ match at offset $d$ by the median correlation of their feature entries.
With $h_{i,a,n}(t)$ the distance to the $n$-th object in $h_{i,a}(t)$, we write this score as
\begin{equation}
m_{ij,ab}(d) = \operatorname*{median}_{n}\ \operatorname{corr}_t\bigl(h_{i,a,n}(t),\, h_{j,b,n}(t+d)\bigr).
\label{eq:instance-score}
\end{equation}
When two cameras see different numbers of other objects, we drop the farthest objects from the longer feature.

As the offset is not yet known, we search jointly, for each pair of cameras $i$ and $j$ and each description $\ell$ that several instances share, over the one-to-one assignment $\pi_{ij,\ell}$ of these instances and the offset $d$,
\begin{equation}
(\pi_{ij,\ell}^\star, d_{ij,\ell}^\star) = \operatorname*{arg\,max}_{\pi_{ij,\ell},\; d\in\mathcal{D}}\ \sum_a m_{ij,a\pi_{ij,\ell}(a)}(d),
\label{eq:instance-search}
\end{equation}
where $\mathcal{D}$ is the set of candidate offsets, from $-50$ to $+50$ frames.
At each candidate offset, we find the best assignment with the Hungarian algorithm~\cite{kuhn1955hungarian}.
We use only the assignment $\pi_{ij,\ell}^\star$, which \cref{sec:identity-results} shows to be accurate.
The offset $d_{ij,\ell}^\star$, however, relies only on the instances of one description, so we drop it and solve for all offsets next, from the cues of all objects.

\subsection{Solving for the offsets}
\label{sec:pairwise-alignment}

With the objects identified across each pair of views, we follow \visualsync and solve \cref{eq:objective} approximately in two steps, which we tailor to our formulation.
We first estimate an offset for each camera pair, and then consolidate them into a single global offset for each camera.

\paragraph{Pairwise synchronization.}
To estimate the offset of cameras $i$ and $j$, we replace the sum over cues in \cref{eq:objective} with robust scores, one for each group of cues.
We split the cues into three groups $g$: the 3D motion ($\mathrm{mot}$), the inter-object distance ($\mathrm{dist}$), and the 2D motion ($\mathrm{2d}$), where the 3D motion holds the displacement and the speed.
As a few extreme values can dominate a correlation, we clip each 3D cue at three median absolute deviations~\cite{hampel1974influence} from its median before correlating it.
To average the correlations within each group into a group score, we then map each correlation through $\operatorname{atanh}$, the Fisher transform~\cite{fisher1921correlation}, and take a trimmed mean~\cite{huber2009robuststatistics},
\begin{equation}
s^{g}_{ij}(d) = \operatorname*{tmean}_{c\in g\cap\mathcal{C}_{ij}}\, \operatorname{atanh}\bigl(r_{ij,c}(d)\bigr),
\label{eq:group-score}
\end{equation}
where $\operatorname{tmean}$ is the mean after discarding the largest 10\% and the smallest 10\% of the values.

To combine the three group scores into one offset, we standardize~\cite{jain2005scorenormalization} each one over the candidate offsets into a $z$-score $z^{g}_{ij}(d)$, as the group scores can have different scales.
We then take the offset at which the weighted sum of the $z$-scores peaks,
\begin{align}
S_{ij}(d) &= z^{\mathrm{mot}}_{ij}(d)+z^{\mathrm{dist}}_{ij}(d)+\lambda_{\mathrm{2d}}\,z^{\mathrm{2d}}_{ij}(d),
\label{eq:fused-score}\\
\hat d_{ij} &= \operatorname*{arg\,max}_{d\in\mathcal{D}}\, S_{ij}(d),
\label{eq:pairwise-offset}
\end{align}
where $\lambda_{\mathrm{2d}}$ is a hyperparameter that controls the influence of the 2D motion.
At each offset, we leave out of this sum any group without a defined correlation.
The offsets of different camera pairs, however, need not agree with one offset per camera.

\paragraph{Global synchronization.}
\label{sec:global-synchronization}
To obtain one offset per camera, we fit the offsets $\boldsymbol{\theta}$ of all cameras to the pairwise offsets $\hat d_{ij}$.
Unlike \visualsync, which discards the pairs with an ambiguous offset, we weight each pair by the peak of $S_{ij}$, \ie, $w_{ij}=\max\bigl(S_{ij}(\hat d_{ij}),0\bigr)$.
As we standardize the group scores over the candidate offsets, this peak is high when the estimated offset stands out from the others.
We also use the Cauchy loss~\cite{barron2019robustloss} instead of the Huber loss of \visualsync, so that a grossly wrong pairwise offset loses its influence on the fit,
\begin{equation}
\hat{\boldsymbol{\theta}} = \operatorname*{arg\,min}_{\boldsymbol{\theta},\,\theta_1=0} \sum_{(i,j)\in\mathcal{E}} w_{ij}\, \rho\left( \theta_i-\theta_j-\hat d_{ij} \right),
\label{eq:global-objective}
\end{equation}
where $\mathcal{E}$ is the set of camera pairs with an estimate $\hat d_{ij}$, and $\rho$ is the Cauchy loss, with its scale set to match a temporal error of one frame.
We initialize the offsets with weighted least squares, and solve \cref{eq:global-objective} with iteratively reweighted least squares~\cite{holland1977irls}.

\section{Experiments}
\label{sec:experiments}

\subsection{Experimental setup}
\label{sec:experimental-setup}

\begin{table*}[t]
\caption{\textbf{Global offset error.}
We report the mean and the median absolute offset error over all cameras, in milliseconds~($\downarrow$), with the standard deviation of our method over the random crops that create the offsets.
The numbers of the other methods are taken from \visualsync.
Methods marked with $^{*}$ use ground-truth camera poses.
Our method assumes calibrated cameras and uses the ground-truth intrinsics.
In Ours w/o wearer, we do not place the wearer at the center of each egocentric camera.
We mark the best value of each column in \textbf{bold} and the second best with an \underline{underline}.
On the three real datasets, our method outperforms every baseline by a large margin.
}
\label{tab:global-offset-estimation}
\vspace{-1em}
\begin{center}
\resizebox{\textwidth}{!}{%
\small
\setlength{\tabcolsep}{9pt}%
\begin{tabular}{@{}lcccccccc@{}}
\toprule
& \multicolumn{2}{c}{EgoHumans}
& \multicolumn{2}{c}{CMU Panoptic}
& \multicolumn{2}{c}{3D-POP}
& \multicolumn{2}{c}{UDBD}\\
\cmidrule(lr){2-3}
\cmidrule(lr){4-5}
\cmidrule(lr){6-7}
\cmidrule(lr){8-9}
Method & Mean & Median & Mean & Median & Mean & Median & Mean & Median\\
\midrule
Uni4D\textsuperscript{*}~\cite{yao2025uni4d}
& 447.4 & 222.1 & 777.9 & 99.9 & 1600.1 & 1265.4 & 103.2 & 25.1\\
\mastr
& 742.3 & 263.8 & 113.4 & 58.1 & 150.3 & \underline{72.2} & 10.1 & 7.4\\
Sync-NeRF\textsuperscript{*}~\cite{kim2024syncnerf}
& - & - & 919.5 & 866.7 & 1138.9 & 1100.0 & \textbf{0.4} & \underline{0.2}\\
\visualsync
& 122.1 & 46.6
& \underline{112.6} & \underline{41.5}
& \underline{114.7} & 77.8 & 20.2 & 5.9\\
Ours
& \textbf{44.5}\,{\footnotesize$\pm4.41$}
& \underline{15.0}\,{\footnotesize$\pm22.36$}
& \textbf{11.1}\,{\footnotesize$\pm3.24$}
& \textbf{0.0}\,{\footnotesize$\pm0.00$}
& \textbf{1.7}\,{\footnotesize$\pm3.73$}
& \textbf{0.0}\,{\footnotesize$\pm0.00$}
& \underline{8.7}\,{\footnotesize$\pm1.48$}
& \textbf{0.0}\,{\footnotesize$\pm0.00$}\\
Ours w/o wearer
& \underline{71.1}\,{\footnotesize$\pm6.04$} & \textbf{5.0}\,{\footnotesize$\pm11.18$} & - & - & - & - & - & -\\
\bottomrule
\end{tabular}%
}
\end{center}
\end{table*}

\vspace{-\customparskip}
\paragraph{Datasets and protocol.}
We evaluate on the four multi-camera datasets that \visualsync uses: \egohumans, \panoptic, \threedpop, and the synthetic Unsynchronized Dynamic Blender Dataset (UDBD)~\cite{kim2024syncnerf}.
We use seven \egohumans scenes with 11--22 cameras each, four \panoptic scenes with 15 cameras each, four \threedpop scenes with 4 cameras each, and three \udbd scenes with 14 cameras each.
Of the 114 \egohumans cameras, 25 are egocentric, \ie, worn by one of the people in the scene.

To create unsynchronized videos, we follow the protocol of \visualsync.
From each scene, we extract a sequence of $\approx$10\,s that all cameras observe.
We then crop $\approx$2--3\,s independently from the beginning and the end of each view, which gives each camera its own offset.
We draw these crops at random with five seeds, and evaluate on all of them.

For all datasets, we use a single set of candidate offsets $\mathcal{D}$, from $-50$ to $+50$ frames, which covers the largest offset that these crops create, $\approx$1\,s.
\visualsync instead sets this range per dataset, \eg, to $\pm30$ frames on \udbd and to $\pm100$ frames on \panoptic.

\paragraph{Implementation details.}
We select our parameters on a \threedpop validation split that is disjoint from the test sequences, and use them for all datasets.
We set $\lambda_{\mathrm{2d}}=0.25$ in \cref{eq:fused-score}.
As in the code of \visualsync, we run at most 10 iterations of iteratively reweighted least squares.
We optimize the offsets in continuous frame units, and round them to integer frames at the end.
Apart from these parameters, we use the pretrained models of \cref{sec:method} without fine-tuning and with their default settings.
For each scene and seed, we prompt \gpt on the unsynchronized videos.

An egocentric camera of \egohumans does not see its wearer, one of the people in the scene.
We thus add the wearer to this camera as one more object, placed at the camera center.
As no description can identify the wearer in the other cameras, we compare the feature of the wearer with that of every person that each other camera sees (\cref{eq:instance-score}), and take the person whose feature agrees best.

\paragraph{Baselines.}
We compare with \visualsync, \syncnerf, and VisualSync's adaptations of \unifourd and \mastr to synchronization.
For all of them, we use the results that \visualsync reports.
We create our crops with the same protocol as \visualsync, but as its crops are not public, ours may differ from them.
We thus report in \cref{tab:global-offset-estimation} the standard deviation of our errors over the five random draws of our crops.
On the three real datasets, the standard deviation of our mean error is much smaller than our lead over every baseline.

We also compare with \sesyn, the pose-based method closest to ours (\cref{sec:related}), which is specific to people.
We run it on \egohumans with its pretrained model and \hrnet poses, without fine-tuning, over our candidate offsets.
To compare the two methods on even ground, we give them the same identities: we evaluate only the camera pairs in which a description identifies the same person in both views.
For each such pair, we run \sesyn on the poses of every person that both views share, and average the offsets that it estimates for these people, as its authors do when several people are in view~\cite[App.~A.2]{yin2022sesyn}.
As \sesyn has no global optimization, we compare it with our pairwise offsets before the global optimization.

\paragraph{Additional information given to each method.}
Our method assumes calibrated cameras, \ie, we give it the ground-truth intrinsics of each camera, but not the camera poses, which it does not need.
\unifourd and \syncnerf instead use ground-truth camera poses~\cite{liu2025visualsync}.
\visualsync estimates both the intrinsics and the poses with \vggt, and knows which cameras move and which are static.
We likewise tell our method which cameras are egocentric, so that it places their wearers.

\paragraph{Metrics.}
As in \visualsync, we report the mean and the median absolute offset error over all cameras, in milliseconds and averaged over the five seeds.
For pairwise offsets, we report their mean absolute error and the area under the cumulative accuracy curve up to 100 and 500\,ms, $A@100$ and $A@500$.

\subsection{Results}
\label{sec:results}

\begin{table}[t]
\caption{\textbf{Comparison with pose-based synchronization on EgoHumans.}
We compare on the camera pairs in which a description identifies the same person in both views, and run the pose-based \sesyn on the \hrnet poses of every person that both views share, averaging the offsets that it estimates for these people.
Mean is the mean absolute error.
We also report the standard deviation of each method over the five seeds of the random crops.
Ours w/o wearer does not place the wearer at the center of each egocentric camera.
With the same identities across views, our pairwise mean error is less than half that of \sesyn.}
\label{tab:pose-based-comparison}
\vspace{-1em}
\begin{center}
\resizebox{\linewidth}{!}{%
\small
\setlength{\tabcolsep}{11pt}%
\begin{tabular}{@{}lccc@{}}
\toprule
Method & Mean (ms) $\downarrow$ & $A@100$ $\uparrow$ & $A@500$ $\uparrow$ \\
\midrule
\sesyn
& 511.1\,{\footnotesize$\pm38.24$}
& 8.1\,{\footnotesize$\pm0.85$}
& 31.1\,{\footnotesize$\pm1.69$} \\
Ours
& \textbf{246.2}\,{\footnotesize$\pm7.81$}
& \textbf{45.8}\,{\footnotesize$\pm0.33$}
& \textbf{70.3}\,{\footnotesize$\pm0.56$} \\
Ours w/o wearer
& 307.0\,{\footnotesize$\pm8.90$}
& 44.0\,{\footnotesize$\pm0.34$}
& 66.7\,{\footnotesize$\pm0.65$} \\
\bottomrule
\end{tabular}%
}
\end{center}
\end{table}

\begin{figure*}[t]
\centering
\begin{subfigure}{0pt}\phantomsubcaption\label{fig:score-curve-ego}\end{subfigure}%
\begin{subfigure}{0pt}\phantomsubcaption\label{fig:score-curve-cmu}\end{subfigure}%
\begin{subfigure}{0pt}\phantomsubcaption\label{fig:score-curve-3dpop}\end{subfigure}%
\begin{subfigure}{0pt}\phantomsubcaption\label{fig:score-curve-udbd}\end{subfigure}%
\includegraphics[width=0.9\textwidth]{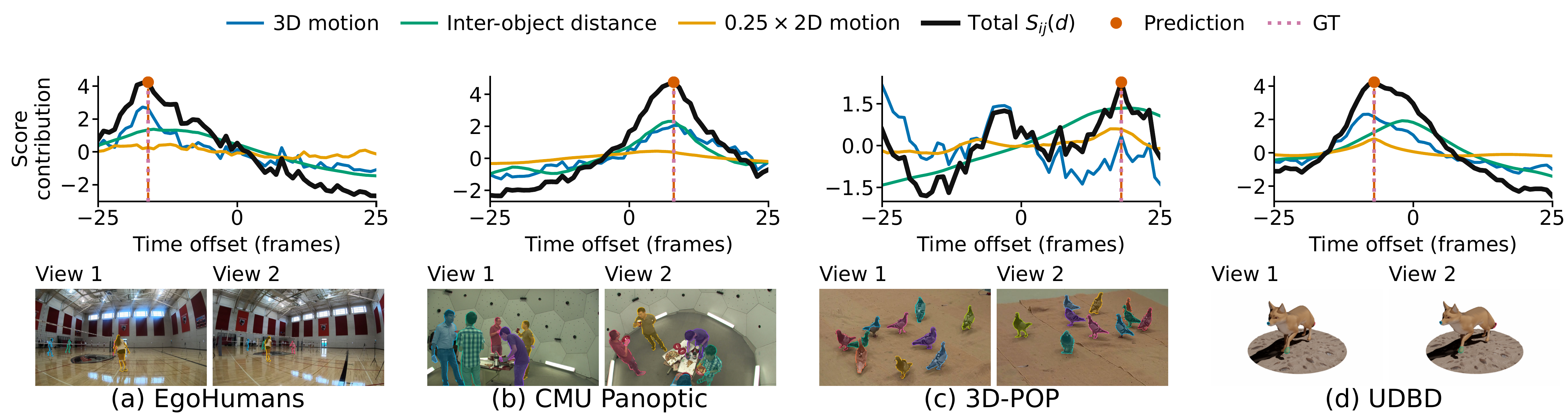}
\caption{\textbf{Example pair alignments with our cues.}
For one camera pair of each dataset, we plot the $z$-scores of the 3D motion and the inter-object distance, $\lambda_{\mathrm{2d}}$ times that of the 2D motion, and their sum $S_{ij}(d)$, over the offsets $d$ with $|d|\le25$ frames.
Below each plot, we show the two views with their segmented objects.
In all four pairs, the peak of the sum, our estimate, coincides with the ground truth.}
\label{fig:pairwise-score-curves}
\end{figure*}

We report the accuracy of the global offsets in \cref{tab:global-offset-estimation}, and show the pairwise scores of our method for one camera pair of each dataset in \cref{fig:pairwise-score-curves}.

\paragraph{Global offset error -- \cref{tab:global-offset-estimation}.}
On the three real datasets, our method outperforms every baseline by a large margin.
On the synthetic \udbd, \syncnerf has near-perfect results, with ground-truth camera poses and a radiance field fitted to each scene.
\syncnerf, however, fails on \panoptic and \threedpop, with mean errors above 900\,ms.
To find the offsets, it aligns a radiance field with every video at the pixel level, which needs pixel-accurate camera poses, such as those of synthetic data.
On real videos, even the ground-truth camera poses have errors, which \syncnerf can confuse with the motion of the scene.
Our method instead compares the relationships between objects, which need neither camera poses nor pixel-level alignment across views.
We believe that this robust signal is why our lead on the real datasets is so large.

Our lead on \egohumans holds even when \visualsync uses ground-truth camera poses.
With ground-truth camera poses, VisualSync has mean and median errors of 72.4 and 28.6\,ms there~\cite[Table~3]{liu2025visualsync}, still above ours.

\paragraph{Pose-based synchronization -- \cref{tab:pose-based-comparison}.}
With the same identities across views, our pairwise mean error on \egohumans before the global optimization is also less than half that of the pose-based \sesyn.
This result suggests that objects and their relationships are a better cue for synchronization than human poses.

\begin{table}[t]
\caption{
\textbf{Contribution of each cue.}
We report the mean and the median global error in milliseconds~($\downarrow$) for the combinations of cues marked with $\checkmark$.
The inter-object distance alone excludes the cameras that see fewer than two objects, where no distance exists.
EgoHumans w/o wearer does not place the wearer at the center of each egocentric camera.
Our full model, in the last row, has the lowest mean error on all four datasets.
All cue types are important in achieving the best results.
}
\label{tab:cue-ablation}
\vspace{-1em}
\begin{center}
\resizebox{\linewidth}{!}{%
\small
\setlength{\tabcolsep}{3pt}%
\renewcommand{\arraystretch}{1.0}%
\begin{tabular}{@{}cccc*{10}{r}@{}}
\toprule
\multicolumn{4}{c}{Cue configuration} & \multicolumn{2}{c}{EgoHumans} & \multicolumn{2}{c}{\shortstack{EgoHumans\\w/o wearer}} & \multicolumn{2}{c}{CMU Panoptic} & \multicolumn{2}{c}{3D-POP} & \multicolumn{2}{c}{UDBD}\\
\cmidrule(lr){1-4}\cmidrule(lr){5-6}\cmidrule(lr){7-8}\cmidrule(lr){9-10}\cmidrule(lr){11-12}\cmidrule(lr){13-14}
\shortstack{3D\\displacement} &
\shortstack{3D\\speed} &
\shortstack{Inter-object\\distance} &
\shortstack{2D\\motion} &
Mean & Median & Mean & Median & Mean & Median & Mean & Median & Mean & Median\\
\midrule
\(\checkmark\) & & & &
333.1 & 70.0 & 333.1 & 70.0 & 377.6 & 186.7 & 521.7 & 493.3 & 24.1 & 3.3\\
& \(\checkmark\) & & &
178.3 & 50.0 & 178.3 & 50.0 & 37.3 & 36.7 & 246.7 & 180.0 & 35.2 & 30.0\\
& & \(\checkmark\) & &
147.7 & 100.0 & 141.7 & 90.0 & 48.9 & 33.3 & 36.7 & 26.7 & 188.6 & 33.3\\
\midrule
\(\checkmark\) & \(\checkmark\) & & &
158.3 & 50.0 & 158.3 & 50.0 & 127.3 & 13.3 & 531.7 & 520.0 & 24.9 & 26.7\\
\(\checkmark\) & & \(\checkmark\) & &
57.3 & 40.0 & 93.4 & 45.0 & 17.8 & 10.0 & 41.7 & \textbf{0.0} & 72.4 & 33.3\\
& \(\checkmark\) & \(\checkmark\) & &
97.7 & 50.0 & 103.5 & 50.0 & 32.0 & 33.3 & 56.7 & 60.0 & 58.3 & 20.0\\
\midrule
\(\checkmark\) & \(\checkmark\) & \(\checkmark\) & &
44.7 & \textbf{0.0} & 71.8 & \textbf{0.0} & 11.3 & \textbf{0.0} & 36.7 & \textbf{0.0} & 60.8 & 33.3\\
& & & \(\checkmark\) &
422.9 & 200.0 & 422.9 & 200.0 & 117.3 & 43.3 & 198.3 & 76.7 & 11.6 & \textbf{0.0}\\
& & \(\checkmark\) & \(\checkmark\) &
111.1 & 70.0 & 129.6 & 60.0 & 36.7 & 36.7 & 30.0 & 30.0 & 130.5 & 23.3\\
\(\checkmark\) & \(\checkmark\) & & \(\checkmark\) &
153.2 & 50.0 & 153.2 & 50.0 & 147.3 & 23.3 & 441.7 & 370.0 & 10.3 & \textbf{0.0}\\
\(\checkmark\) & & \(\checkmark\) & \(\checkmark\) &
50.3 & 30.0 & 78.5 & 40.0 & 17.8 & 3.3 & 6.7 & \textbf{0.0} & 9.4 & \textbf{0.0}\\
& \(\checkmark\) & \(\checkmark\) & \(\checkmark\) &
87.6 & 50.0 & 99.7 & 50.0 & 26.0 & 26.7 & 31.7 & 23.3 & 9.1 & \textbf{0.0}\\
\(\checkmark\) & \(\checkmark\) & \(\checkmark\) & \(\checkmark\) &
\textbf{44.5} & 15.0 & \textbf{71.1} & 5.0 & \textbf{11.1} & \textbf{0.0} & \textbf{1.7} & \textbf{0.0} & \textbf{8.7} & \textbf{0.0}\\
\bottomrule
\end{tabular}%
}
\end{center}
\end{table}

\subsection{Ablation study}
\label{sec:ablation}

To motivate our design choices, we evaluate our method with its components removed or replaced.

\paragraph{Cues -- \cref{tab:cue-ablation,fig:pairwise-score-curves}.}
\label{sec:cue-ablation}
Each cue alone has a large error on at least one dataset, but all four together have the lowest mean error on every dataset.
Removing the inter-object distance from the full model raises the mean error the most on the three real datasets,
but not on the synthetic \udbd, where the `objects' are often parts of a single animal, whose relationships do not change much.
This, however, is not the case on the real datasets.

As shown in \cref{fig:pairwise-score-curves}, even on a single camera pair, the cues are more reliable together than alone.
Each cue alone can be noisy or peak away from the ground-truth offset, as the 3D motion and the inter-object distance do on \threedpop in \cref{fig:score-curve-3dpop}.
Their sum, however, peaks at the ground-truth offset in all four pairs.

\paragraph{Wearer at the camera center.}
Placing the wearer at the camera center adds an additional object (person) to the egocentric cameras of \egohumans, and halves their mean error, from 249.2 to 126.4\,ms.
It slightly raises that of the static cameras, from 21.1 to 21.5\,ms.
Such a large gain from a single added object, for which we only need to know which cameras are egocentric, shows how strong a cue the relationships between objects are.
Even without the wearer, our method still outperforms every baseline on \egohumans in \cref{tab:global-offset-estimation}.

\paragraph{Object identities across views.}
\label{sec:identity-results}
Our pairwise offsets rely on our instance matching, which pairs the instances across views accurately, even before it knows the offsets.
On \threedpop, where the VLM cannot tell the pigeons apart and gives them all one description, 84\% of the pigeon pairs that it matches are correct, \ie, show the same individual pigeon in both views.
For the egocentric cameras of \egohumans with a ground-truth wearer, 97.3\% of the wearer matches in the other cameras are correct.

\begin{table}[t]
\caption{\textbf{Contribution of the identities from the VLM.}
We report the mean and the median global error in milliseconds~($\downarrow$), and the percentage of the matches across views that are correct (Corr.,~$\uparrow$).
In Ours w/o identities, we use the VLM only to find the objects in each view, and match all objects across views with our instance matching.
In Ours w/o wearer, we do not place the wearer at the center of each egocentric camera.
Our full method has the lowest mean error on the three real datasets, and the results change little on UDBD, where it is mostly a single synthetic object.}
\label{tab:vlm-identities}
\vspace{-1em}
\begin{center}
\resizebox{\linewidth}{!}{%
\small
\setlength{\tabcolsep}{3pt}%
\renewcommand{\arraystretch}{1.0}%
\begin{tabular}{@{}l*{11}{r}@{}}
\toprule
& \multicolumn{3}{c}{EgoHumans} & \multicolumn{3}{c}{CMU Panoptic} & \multicolumn{3}{c}{3D-POP} & \multicolumn{2}{c}{UDBD}\\
\cmidrule(lr){2-4}\cmidrule(lr){5-7}\cmidrule(lr){8-10}\cmidrule(lr){11-12}
Method & Mean & Median & Corr. & Mean & Median & Corr. & Mean & Median & Corr. & Mean & Median\\
\midrule
Ours & 44.5 & 15.0 & 99.2 & 11.1 & 0.0 & 99.9 & 1.7 & 0.0 & 84.0 & 8.7 & 0.0\\
Ours w/o identities & 96.0 & 50.0 & 70.2 & 28.7 & 30.0 & 72.4 & 1.7 & 0.0 & 84.0 & 8.4 & 0.0\\
Ours w/o wearer & 71.1 & 5.0 & 99.2 & - & - & - & - & - & - & - & -\\
Ours w/o wearer and identities & 128.0 & 50.0 & 73.8 & - & - & - & - & - & - & - & -\\
\bottomrule
\end{tabular}%
}
\end{center}
\end{table}

\paragraph{Identities from the VLM -- \cref{tab:vlm-identities}.}
To test how much the identities from the VLM matter, we use the VLM only to find the objects in each view, and match all objects across views with the instance matching of \cref{eq:instance-search}.
Without the identities, fewer matches are correct, and our error rises on \egohumans and \panoptic.
On \threedpop, where the VLM already gives all pigeons one description, the results do not change.
On \udbd, they change little as it is mostly about a single synthetic object.

\paragraph{Offset of the instance matching.}
The offset $d_{ij,\ell}^\star$ at which our instance matching pairs the instances is less accurate than our pairwise offsets.
On \threedpop, its $A@100$ and $A@500$ are 30.6 and 76.7, against 97.8 and 99.6 for our pairwise offsets, even before the global optimization.
Matching the identities across views is thus not enough to synchronize the cameras.
For accurate offsets, we need to compare the cues of all objects, which these identities make possible.

\paragraph{Confidence weights and Cauchy loss.}
Both our confidence weights $w_{ij}$ and the Cauchy loss, which we use instead of the Huber loss of \visualsync, lower the global mean error on the \threedpop test sequences.
With uniform weights and the Huber loss, the global mean error is 10.0\,ms.
The confidence weights alone lower it to 8.3\,ms, the Cauchy loss alone to 5.0\,ms, and both together to 1.7\,ms.
See the \supp for other settings of the global solver.

\paragraph{Further ablations.}
We further ablate the descriptions and each coordinate of the displacement, test another video depth model, and report the accuracy of the pairwise offsets and the runtime of our method.
We also sweep the weights of the 2D motion, the displacement, and the inter-object distance, and the settings of the point tracking and the scoring.
Please see the \supp for these.

\section{Conclusion}
\label{sec:conclusion}

We have proposed a method to synchronize videos of the same dynamic event through the relationships between objects.
In each camera independently, we measure the 3D motion of the objects and the distances between them, and compare these cues across cameras through the identities of the objects, which we obtain with foundation models.
We then estimate the offset of each camera pair, and combine the pairwise offsets into one consistent offset per camera, weighting each pair by its confidence.
We have demonstrated that our method outperforms the state of the art by a large margin on real-world videos.
Our ablation study further demonstrates the importance of the relationships between objects, and of combining all cues.

\paragraph{Limitations and future work.}
Our method currently assumes calibrated cameras, \ie, known intrinsics.
A promising direction is to lift this assumption, which should be possible with off-the-shelf calibration methods, or with the intrinsics that geometry foundation models such as \vggt estimate from the videos themselves, as \visualsync does.
As all our cues come from pretrained foundation models, our method should also improve as these models do, without any training of its own.
Another limitation is that our current framework is discrete, as it estimates the offsets at integer frames, for videos of one frame rate.
A continuous formulation should be possible, and would allow sub-frame accuracy, as well as videos of different frame rates.
More broadly, the relationships between objects could serve as cues for other global reasoning about cameras, such as their calibration and localization.

\section*{Acknowledgments}
This work was supported in part by the Natural Sciences and Engineering Research Council of Canada (NSERC) Discovery Grant, Digital Research Alliance of Canada, and Advanced Research Computing at the University of British Columbia.
This research was also supported by the MSIT(Ministry of Science and ICT), Korea, under the Global Research Support Program in the Digital Field program(IITP-2026-RS-2024-00418641) and the ITRC(Information Technology Research Center) support program(IITP-2026-RS-2024-00437102) supervised by the IITP(Institute for Information \& Communications Technology Planning \& Evaluation).

{
    \small
    \bibliographystyle{ieeenat_fullname}
    \bibliography{main}
}

\appendix
\clearpage
\maketitlesupplementary

\section{Pairwise offset accuracy}
\label{sec:supp-pairwise}

\begin{table*}[t]
\caption{\textbf{Pairwise offset accuracy.}
We report the normalized area under the cumulative accuracy curve of the pairwise offsets up to an error of 100 or 500\,ms, $A@100$ and $A@500$, in percent~($\uparrow$).
We give the standard deviations of our method over the random crops that create the offsets, in percentage points.
For the other methods, we give the results that \visualsync reports, which include no result of \syncnerf on \egohumans.
Methods marked with $^{*}$ use ground-truth camera poses, and our method uses the ground-truth intrinsics.
Ours w/o wearer does not place the wearer at the center of each egocentric camera of \egohumans.
We mark the best value of each column in \textbf{bold} and the second best with an \underline{underline}.
Before the global optimization, our pairwise offsets are the most accurate on the three real datasets.}
\label{tab:pairwise-offset-accuracy}
\vspace{-1em}
\begin{center}
\resizebox{\textwidth}{!}{%
\small
\setlength{\tabcolsep}{8pt}%
\begin{tabular}{@{}lcccccccc@{}}
\toprule
& \multicolumn{2}{c}{EgoHumans}
& \multicolumn{2}{c}{CMU Panoptic}
& \multicolumn{2}{c}{3D-POP}
& \multicolumn{2}{c}{UDBD}\\
\cmidrule(lr){2-3}
\cmidrule(lr){4-5}
\cmidrule(lr){6-7}
\cmidrule(lr){8-9}
Method & \(A@100\) & \(A@500\)
& \(A@100\) & \(A@500\)
& \(A@100\) & \(A@500\)
& \(A@100\) & \(A@500\)\\
\midrule
Uni4D\textsuperscript{*}~\cite{yao2025uni4d}
& 23.8 & 49.4 & \underline{32.3} & \underline{60.7}
& 0.9 & 9.5 & 46.2 & 74.1\\
\mastr
& 24.3 & 50.4 & 29.6 & 49.8
& 15.7 & 69.1 & 77.8 & 95.4\\
Sync-NeRF\textsuperscript{*}~\cite{kim2024syncnerf}
& - & - & 3.0 & 13.8
& 0.0 & 8.2 & \textbf{86.7} & \textbf{97.4}\\
\visualsync
& 33.9 & 55.8
& 26.0 & 51.2
& \underline{33.3} & \underline{69.3}
& 82.1 & 94.3\\
Ours
& \textbf{45.8}\,{\footnotesize$\pm0.33$}
& \textbf{70.3}\,{\footnotesize$\pm0.56$}
& \textbf{68.2}\,{\footnotesize$\pm0.56$}
& \textbf{90.6}\,{\footnotesize$\pm0.76$}
& \textbf{97.8}\,{\footnotesize$\pm2.32$}
& \textbf{99.6}\,{\footnotesize$\pm0.47$}
& \underline{83.9}\,{\footnotesize$\pm2.64$}
& \underline{95.5}\,{\footnotesize$\pm1.26$}\\
Ours w/o wearer
& \underline{44.0}\,{\footnotesize$\pm0.34$} & \underline{66.7}\,{\footnotesize$\pm0.65$} & - & - & - & - & - & -\\
\bottomrule
\end{tabular}%
}
\end{center}
\end{table*}

As in \visualsync, we evaluate the pairwise offsets before the global optimization.
In \cref{tab:pairwise-offset-accuracy}, our pairwise offsets are the most accurate of all methods on the three real datasets.
On \panoptic, our $A@100$ is more than twice that of \visualsync.
As both sets of pairwise offsets precede the global optimization, our lead does not come from our confidence weights or the Cauchy loss.

\section{Additional cue ablation}

To see which coordinate gives the gain from the displacement, we add one coordinate at a time to the speed and the inter-object distance in \cref{tab:displacement-coordinates}.
As in the rows with only 3D cues in \cref{tab:cue-ablation} of the main paper, we leave out the 2D motion, so that it does not mask the effect of the coordinates.
On \threedpop, the vertical coordinate $y$ or the depth coordinate $z$ alone gives the gain from the displacement, and on \panoptic, $z$ does.
On \egohumans, $y$ gives most of the gain, but all three coordinates together give the lowest error.
Without the wearer, $y$ alone gives the gain.
On \udbd, adding all three coordinates does not lower the error when we leave out the 2D motion.

We correlate the signed coordinates, so a coordinate can correlate negatively at the true offset, as $x$ and $z$ do for cameras that face each other.
Replacing $x$ and $z$ with the speed in the $xz$-plane, or, for each camera pair, flipping the sign of each coordinate whose strongest correlation over all candidate offsets is negative, does not lower the mean error on the three real datasets.
We use all three coordinates, as we do not assume a particular camera orientation.

\begin{table}[t]
\caption{\textbf{Displacement coordinates.}
We report the global mean error in milliseconds~($\downarrow$) without the 2D motion.
Each row adds the listed coordinates of the displacement, the horizontal $x$, the vertical $y$, and the depth $z$, to the speed and the inter-object distance.
EgoHumans w/o wearer does not place the wearer at the center of each egocentric camera.
Which coordinate lowers the error the most depends on the dataset.}
\label{tab:displacement-coordinates}
\vspace{-1em}
\begin{center}
\resizebox{\linewidth}{!}{%
\footnotesize
\setlength{\tabcolsep}{5pt}%
\begin{tabular}{@{}lccccc@{}}
\toprule
Added coordinates & EgoHumans & \shortstack{EgoHumans\\w/o wearer} & CMU Panoptic & 3D-POP & UDBD\\
\midrule
None & \phantom{0}97.72 & 103.51 & 32.00 & \phantom{0}56.67 & 58.25\\
$x$ & 111.67 & 122.11 & 30.89 & 236.67 & 75.56\\
$y$ & \phantom{0}49.91 & \phantom{0}62.46 & 34.00 & \phantom{00}5.00 & 55.40\\
$z$ & \phantom{0}71.58 & \phantom{0}84.56 & \phantom{0}8.44 & \phantom{00}6.67 & 76.03\\
$x$, $y$, $z$ & \phantom{0}44.74 & \phantom{0}71.84 & 11.33 & \phantom{0}36.67 & 60.79\\
\bottomrule
\end{tabular}%
}
\end{center}
\end{table}

\section{Perception inputs}
\label{sec:supp-perception}

We test how much our results depend on the descriptions and on the depth estimator.

\paragraph{Sensitivity to the descriptions.}
Manually written descriptions improve our results only slightly over those of \gpt.
We compare the descriptions of \gpt with manually written ones on one \panoptic scene outside our test scenes.
The manually written descriptions are more detailed, \eg, ``man in a pink checkered shirt and blue jeans'' instead of ``man in a pink checkered shirt'' from \gpt.
With the manual descriptions, the global mean error goes from 12.89 to 10.67\,ms, and the pairwise mean error from 46.60 to 45.56\,ms.

\paragraph{Depth estimator.}
Another depth estimator raises our global mean error, mostly on the egocentric cameras of \egohumans.
We replace metric \vda with \velodepth, another recent metric video depth estimator, while keeping the other settings fixed.
Averaged over the four datasets, the global mean error rises from 16.50 to 30.89\,ms, still below that of \visualsync in \cref{tab:global-offset-estimation}.
The pairwise mean error, averaged in the same way, rises from 84.59 to 121.72\,ms.
When we synchronize \egohumans without its egocentric cameras, the global mean error averaged over the four datasets rises from 11.11 to 15.85\,ms.
On the egocentric cameras alone, the mean error rises from 126.40 to 288.00\,ms.

\section{Implementation details}
\label{sec:supp-implementation}

We give the equation and settings that the main paper describes in words.

\paragraph{2D motion.}
Let $\mathbf{x}_{i,k}(t)$ be the position of track $k$ in camera $i$, and $\Delta\mathbf{x}_{i,k}(t) = \mathbf{x}_{i,k}(t) - \mathbf{x}_{i,k}(t-1)$ its displacement.
Let $\mathcal{T}_i(t)$ be the tracks that are visible in at least 75\% of the frames and in frames $t-1$ and $t$.
The 2D motion is the mean magnitude of the valid track displacements,
\begin{equation}
v^{\mathrm{2D}}_i(t) =
\frac{1}{|\mathcal{T}_i(t)|}
\sum_{k\in\mathcal{T}_i(t)}
\left\lVert \Delta\mathbf{x}_{i,k}(t) \right\rVert_2.
\label{eq:tracker-curve}
\end{equation}
We define $v^{\mathrm{2D}}_i(t)$ only when $|\mathcal{T}_i(t)|>0$.

\paragraph{Point tracking.}
We query 64 points, the 32 strongest SIFT keypoints~\cite{lowe2004sift} in the first frame and the 32 strongest in the middle frame.

\paragraph{Depth scale over time.}
We run metric \vda with its default settings, which process each video in overlapping 32-frame windows.
The depth scale of metric \vda changes little from one window to the next.
On the frames that two consecutive windows share, the depth ratio between the windows deviates from one by 1.8 to 2.8\% on average, depending on the dataset.
On the egocentric cameras of \egohumans alone, the deviation is 3.5\%.

\paragraph{Instance matching.}

For the wearer of an egocentric camera, we search jointly over the person instances of each other camera and the candidate offsets, as in \cref{eq:instance-search}.
When the wearer is out of view of the other camera, we can still match it wrongly to the person with the highest score.

To evaluate our instance matching, we count a match as correct when both instances show the same object, which we find from the ground-truth keypoints that fall in each mask.
The egocentric cameras with a ground-truth wearer are those whose wearer the dataset annotates, except one that its official configuration marks as invalid.

\section{Parameter sweeps}
\label{sec:parameter-ablation}

\Cref{tab:parameter-ablation} reports the effect of each choice on the \threedpop test sequences.
In the weight sweeps, we weight the scores of the displacement cues by $\beta_{\mathrm{disp}}$ after trimming, and scale the standardized distance-group score by $\lambda_{\mathrm{dist}}$.
When $\beta_{\mathrm{disp}}=0$, we exclude the displacement cues before trimming.

\begin{table*}[t]
\caption{\textbf{Parameter sweeps on 3D-POP.}
We evaluate on the \threedpop test sequences and aggregate the results over five random seeds.
Each entry is the change from our setting ($\checkmark$), which we select on the validation split.
Accuracies are in percentage points (pp).
$A@100$ and $A@500$ are the normalized areas under the cumulative accuracy curve of the pairwise offsets up to an error of 100 and 500\,ms.
Only a query budget of 128 and $\lambda_{\mathrm{2d}}=0.10$ lower the global mean error.}
\label{tab:parameter-ablation}
\vspace{-1em}
\begin{center}
\resizebox{\textwidth}{!}{%
\footnotesize
\setlength{\tabcolsep}{9pt}%
\renewcommand{\arraystretch}{0.90}%
\begin{tabular}{@{}llrrrrrr@{}}
\toprule
& & \multicolumn{2}{c}{$\Delta$Global error (ms) $\downarrow$} & \multicolumn{2}{c}{$\Delta$Pairwise error (ms) $\downarrow$} & \multicolumn{2}{c}{$\Delta$Pairwise accuracy (pp) $\uparrow$}\\
\cmidrule(lr){3-4}\cmidrule(lr){5-6}\cmidrule(lr){7-8}
Category & Setting & Mean & Median & Mean & Median & $A@100$ & $A@500$\\
\midrule
Query budget & 32 & $+36.67$ & $+3.33$ & $+18.89$ & 0.00 & $-2.22$ & $-3.11$\\
 & 64 $\checkmark$ & 0.00 & 0.00 & 0.00 & 0.00 & 0.00 & 0.00\\
 & 128 & $-1.67$ & 0.00 & 0.00 & 0.00 & 0.00 & 0.00\\
\midrule
Query allocation & 48/16 & $+1.67$ & 0.00 & 0.00 & 0.00 & 0.00 & 0.00\\
 & 32/32 $\checkmark$ & 0.00 & 0.00 & 0.00 & 0.00 & 0.00 & 0.00\\
 & 16/48 & $+1.67$ & 0.00 & 0.00 & 0.00 & 0.00 & 0.00\\
\midrule
Track persistence & 0.50 & $+5.00$ & 0.00 & $+24.44$ & $+6.67$ & $-3.33$ & $-3.33$\\
 & 0.75 $\checkmark$ & 0.00 & 0.00 & 0.00 & 0.00 & 0.00 & 0.00\\
\midrule
Group-score standardization & None & $+1.67$ & 0.00 & $-1.11$ & 0.00 & $+0.56$ & $+0.11$\\
 & Standard $z$-score $\checkmark$ & 0.00 & 0.00 & 0.00 & 0.00 & 0.00 & 0.00\\
\midrule
2D motion weight $\lambda_{\mathrm{2d}}$ & 0 & $+35.00$ & 0.00 & $+23.33$ & 0.00 & $-2.78$ & $-3.22$\\
 & 0.10 & $-1.67$ & 0.00 & 0.00 & 0.00 & 0.00 & 0.00\\
 & 0.25 $\checkmark$ & 0.00 & 0.00 & 0.00 & 0.00 & 0.00 & 0.00\\
 & 0.50 & $+1.67$ & 0.00 & $+24.44$ & $+6.67$ & $-3.33$ & $-3.33$\\
 & 1.00 & $+1.67$ & 0.00 & $+33.33$ & $+3.33$ & $-6.11$ & $-4.56$\\
\midrule
Displacement weight $\beta_{\mathrm{disp}}$ & 0 & $+30.00$ & $+23.33$ & $+70.00$ & $+46.67$ & $-45.56$ & $-11.67$\\
 & 0.25 & $+10.00$ & $+6.67$ & $+14.44$ & $+13.33$ & $-13.33$ & $-2.44$\\
 & 0.50 & $+3.33$ & 0.00 & $+5.56$ & $+3.33$ & $-4.44$ & $-0.89$\\
 & 1.00 $\checkmark$ & 0.00 & 0.00 & 0.00 & 0.00 & 0.00 & 0.00\\
 & 2.00 & $+1.67$ & 0.00 & 0.00 & 0.00 & 0.00 & 0.00\\
\midrule
Distance-group weight $\lambda_{\mathrm{dist}}$ & 0.50 & $+35.00$ & $+3.33$ & $+285.56$ & $+150.00$ & $-28.89$ & $-29.78$\\
 & 1.00 $\checkmark$ & 0.00 & 0.00 & 0.00 & 0.00 & 0.00 & 0.00\\
 & 2.00 & 0.00 & 0.00 & 0.00 & 0.00 & 0.00 & 0.00\\
\midrule
Correlation transform & Fisher transform $\checkmark$ & 0.00 & 0.00 & 0.00 & 0.00 & 0.00 & 0.00\\
 & None & $+33.33$ & $+3.33$ & $+65.56$ & $+63.33$ & $-7.78$ & $-9.56$\\
\midrule
3D clipping / score aggregation & No clipping, 10\% trimmed & $+1.67$ & 0.00 & $-1.11$ & 0.00 & $+0.56$ & $+0.11$\\
 & $\pm3$, 10\% trimmed $\checkmark$ & 0.00 & 0.00 & 0.00 & 0.00 & 0.00 & 0.00\\
 & $\pm3$, ordinary mean & $+6.67$ & 0.00 & $+50.00$ & $+10.00$ & $-7.78$ & $-6.89$\\
\midrule
Global solver & Uniform + least squares & $+25.00$ & $+20.00$ & 0.00 & 0.00 & 0.00 & 0.00\\
 & Confidence + least squares & $+21.67$ & $+16.67$ & 0.00 & 0.00 & 0.00 & 0.00\\
 & Uniform + Huber & $+8.33$ & $+3.33$ & 0.00 & 0.00 & 0.00 & 0.00\\
 & Confidence + Huber & $+6.67$ & $+3.33$ & 0.00 & 0.00 & 0.00 & 0.00\\
 & Uniform + Cauchy & $+3.33$ & 0.00 & 0.00 & 0.00 & 0.00 & 0.00\\
 & Confidence + Cauchy $\checkmark$ & 0.00 & 0.00 & 0.00 & 0.00 & 0.00 & 0.00\\
\bottomrule
\end{tabular}%
}
\end{center}
\end{table*}

On the \threedpop test sequences, removing group-score standardization or the clipping of the 3D cues slightly improves pairwise accuracy but raises the global mean error, the metric by which we select parameters on the validation split.
A query budget of 128 and $\lambda_{\mathrm{2d}}=0.10$ give slightly lower global mean errors, and $\lambda_{\mathrm{dist}}=2$ changes no metric.
We still keep the settings selected on the validation split.

\section{Runtime}

Our full pipeline takes 1.3\,h for 15 \panoptic videos of 200 frames each on one NVIDIA RTX 5090 GPU.
\visualsync reports 3.3\,h for the same input size on one NVIDIA A6000 GPU.

\section{Identifying the objects}
\label{sec:supp-instruction}

We obtain the identities from \gpt with the instruction in \cref{fig:supp-instruction}.
For each object, we give its description, the short noun phrase in its \texttt{prompt} field, to \samthree as the text prompt.

\begin{figure*}[tp]
\begin{tcolorbox}[
    enhanced,
    colback=black!2,
    colframe=black!30,
    boxrule=0.35pt,
    arc=0pt,
    left=4pt,
    right=4pt,
    top=3pt,
    bottom=3pt
]
\begin{Verbatim}[
    breaklines=true,
    breakanywhere=true,
    breaksymbolleft={},
    breaksymbolright={},
    fontsize=\footnotesize
]
You select segmentation targets for constructing temporal object trajectories to synchronize an unsynchronized multi-camera scene.
All input images belong to the same scene.
Each camera sheet contains three samples in local chronological order.
Infer temporal changes only within each camera.
Across cameras, compare semantic identity and visibility without assuming temporal simultaneity.
Perform the following three stages before producing the final output.

STAGE 1 - PER-CAMERA DISCOVERY
Independently examine each camera and identify candidate foreground entities and components relevant to the scene's primary interactions.
1. Include focal entities, manipulated or independently moving objects, and interaction-related components.
2. For articulated, deformable, or multipart entities, consider a coarse-to-fine hierarchy from the whole entity to meaningful components and localized subcomponents.
3. Retain components only when they are visibly supported, semantically coherent, independently segmentable, and potentially informative through motion relative to their parent.
4. Examine fine-grained components, distinct functional parts, and repeated nearby objects.
Do not invent unseen components or divide objects into arbitrary appearance-based fragments.
5. Consider at most one localized interaction or support region when it provides a useful temporal reference.

STAGE 2 - CROSS-CAMERA CONSOLIDATION
6. Merge equivalent descriptions of the same physical entity or component across cameras while preserving distinct instances.
7. Prefer candidates supported by multiple cameras, but retain a clearly observed component from a single camera when its parent is supported across views.
8. Use stable semantic and visual characteristics for cross-camera identification.
Never assume that observations from different cameras are simultaneous.
9. For repeated or ambiguous instances of the same category, use instance_policy "separate_then_match" to preserve individual instances for downstream association.

STAGE 3 - TRAJECTORY-ORIENTED SELECTION
10. Prioritize focal entities and components that provide distinctive, reliable temporal trajectories.
11. Retain complete entities and their components together only when they provide meaningfully different motion information.
12. For articulated or deformable entities, prefer the smallest coherent components that capture independent relative motion.
Preserve multiple independently moving components rather than replacing them with a redundant parent region.
13. Retain manipulated objects, distinct functional components, and repeated neighboring entities when they contribute complementary motion or interaction information.
14. Exclude irrelevant background regions, incidental entities, unstable fragments, and components with redundant or unreliable trajectories.
Retain secondary entities only when they provide meaningful temporal cues.
15. Select at most 16 targets, prioritizing focal importance, independent motion, cross-camera support, segmentation reliability, and non-redundancy.
16. Verify that no supported focal entity or independently informative component has been omitted.

OUTPUT
Return only valid JSON without explanations or intermediate results.
Each target must represent one coherent, segmentable region described by a concise singular English noun phrase.
Use lowercase snake_case names and consecutive global_id values starting from 1.
Use instance_policy "single" for a uniquely identifiable instance and "separate_then_match" for repeated or ambiguous instances requiring separate segmentation and later association.
{ "objects": [ { "global_id": 1, "name": "stable_snake_case_name", "prompt": "short English segmentation prompt", "instance_policy": "single" } ] }
\end{Verbatim}
\end{tcolorbox}
\caption{\textbf{Instruction for identifying the objects.}
We give \gpt this instruction once, with the first, middle, and last frames of every camera.}
\label{fig:supp-instruction}
\end{figure*}

\end{document}